\documentclass[letterpaper, 10 pt, conference]{ieeeconf}  % Comment this line out if you need a4paper

\IEEEoverridecommandlockouts                              % This command is only needed if 
\usepackage{color, colortbl}
\usepackage[dvipsnames]{xcolor}
\usepackage{amssymb}
\usepackage{stix}
\usepackage{amssymb,amsmath}
\usepackage{booktabs}
\usepackage{algpseudocode}
\usepackage{algorithm}
\definecolor{LGreen}{rgb}{0.89,1,0.89}
\definecolor{DGreen}{rgb}{0,0.59,0}
\definecolor{Gray}{gray}{0.9}
\usepackage{multirow}
\usepackage{mathtools}
\usepackage{xspace}
\usepackage{graphicx}
\usepackage{afterpage}
\usepackage{pifont}% http://ctan.org/pkg/pifont
\newcommand{\xmark}{\ding{55}}%
\usepackage{mathtools}
\usepackage{authblk}
\def\ie{i.e.\@\xspace}

\title{\LARGE \bf
Uplifting Range-View-based 3D Semantic Segmentation in Real-Time with Multi-Sensor Fusion
}

\author{Shiqi Tan$^{1,2,*}$, Hamidreza Fazlali$^{2,*}$, Yixuan Xu$^{1,2}$, Yuan Ren$^{2}$, and Bingbing Liu$^{2}$% <-this % stops a space
\\
  $^{1}$University of Toronto \\
  $^{2}$Huawei Noah's Ark Lab \\
  \texttt{\{alexshiqi.tan, richardyixuan.xu\}@mail.utoronto.ca} \\ \texttt{\{hamidreza.fazlali1, yuan.ren3, liu.bingbing\}@huawei.com} 
\thanks{$^{*}$ indicates equal contribution.}%
}

\begin{document}

\maketitle
\thispagestyle{empty}
\pagestyle{empty}

%%%%%%%%%%%%%%%%%%%%%%%%%%%%%%%%%%%%%%%%%%%%%%%%%%%%%%%%%%%%%%%%%%%%%%%%%%%%%%%%
\begin{abstract}

Range-View(RV)-based 3D point cloud segmentation is widely adopted due to its compact data form. However, RV-based methods fall short in providing robust segmentation for the occluded points and suffer from distortion of projected RGB images due to the sparse nature of 3D point clouds. To alleviate these problems, we propose a new \textcolor{black}{\textbf{L}iDAR \textbf{a}nd \textbf{C}amera \textbf{Range}-view-based 3D point cloud semantic segmentation method (LaCRange)}. Specifically, a distortion-compensating knowledge distillation (DCKD) strategy is designed to remedy the adverse effect of RV projection of RGB images. Moreover, a context-based feature fusion module is introduced for robust and preservative sensor fusion. Finally, in order to address the limited resolution of RV and its insufficiency of 3D topology, a new point refinement scheme is devised for proper aggregation of features in 2D and augmentation of point features in 3D. We evaluated the proposed method on large-scale autonomous driving datasets \ie SemanticKITTI and nuScenes. In addition to being real-time, the proposed method achieves state-of-the-art results on nuScenes benchmark.

\end{abstract}

%%%%%%%%%%%%%%%%%%%%%%%%%%%%%%%%%%%%%%%%%%%%%%%%%%%%%%%%%%%%%%%%%%%%%%%%%%%%%%%%
\section{INTRODUCTION}

Semantic segmentation is a core computer vision task in the perception systems of autonomous vehicles (AVs). This detailed semantic information needs to be acquired in real-time to enable the AVs to better understand their surroundings to make safer decisions. 

Outdoor 3D semantic segmentation has been receiving increasing attention due to the availability of public datasets such as nuScenes \cite{nuscenes} and SemanticKITTI \cite{semantickitti}. For better and more robust scene understanding, AVs are usually equipped with multiple types of sensors such as camera and LiDAR. Camera images have dense pixel representation that provides rich textural information. However, cameras face inherent challenges under adverse weather conditions and poor illumination. On the contrary, LiDAR sensors are less susceptible to lighting conditions, whilst providing accurate geometric and distance information.

% \begin{figure}[!t]
%   \centering
%   \includegraphics[width=1\linewidth, height=5cm]{First_Page.pdf}
%    \caption{Segmentation performance of LaCRange with DCKD for addressing distortion in RV-RGB. From top to bottom: Original RGB, RV-RGB, Student (scratch) prediction, Student (DCKD) prediction, Ground-truth semantics. Best viewed in color.}
%    \label{fig:figure1}
%    \vspace{-6mm}
% \end{figure}

For alleviating weaknesses and leveraging complementary information of each sensor, fusion-based strategies were proposed. One approach for multi-sensor feature fusion is to project the image features to 3D LiDAR coordinates using spherical (Range-View (RV)) projection to augment corresponding point features \cite{rgbal}. Despite the efficiency of RV-based methods due to compact representation, most of the appearance information is lost due to the sparse nature of 3D point clouds. Recently, PMF \cite{pmf} exploits perspective projection to project points onto the camera coordinates to preserve appearance information. Nevertheless, this approach comes with large computational complexity when multiple cameras are involved. The trade-off between information preservation and computational cost poses a non-trivial yet imperative challenge for the fusion of camera and LiDAR in real-time semantic segmentation.

Additionally, as RV-based methods only process a subset of the input point cloud, prior works usually resort to k-nearest-neighbors (kNN) post-processing technique to predict a label for every 3D point \cite{rangenet++, salsanext}. This imposes a large gap between segmentation performance on RV-projected points and the entire point cloud. % one word

In this paper, we seek to extend the capability of RV-based methods while still preserving its low latency advantage. We propose \textbf{LaCRange}, a new \textbf{L}iDAR \textbf{a}nd \textbf{C}amera \textbf{Range}-view-based semantic segmentation method for 3D point clouds.
%by proposing \textcolor{red}{LacRange, a new Lidar and camera Range-view-based semantic segmentation network for 3D point cloud.} 
To address the appearance information loss caused by RV projection of RGB images, we devise a distortion-compensating knowledge distillation (DCKD) strategy by using a teacher model, which is pre-trained to learn how to extract detailed and structurally intact information from original images. Compared to prior works \cite{pmf, rgbal}, our image encoder is lightweight yet powerful since it is guided by the teacher.
For fusing the camera and LiDAR features, a new context-based feature fusion module is proposed. In this strategy, the camera and LiDAR features are concatenated and processed to generate global context information of these features. Then, their global context information interact with each other through Modality Look-Up (MLU), which retrieve information from each modality to supplement the initially fused features.

Moreover, in order to further close the segmentation performance gap between the projected points and the full point cloud, a new point refinement scheme is proposed. In this method, a Semantic-Range-Remission Feature Aggregation (SR$^{2}$FA) module with a dynamic kernel is proposed, which aggregates features in each pixel of the RV image that has a projected point. Then, a 3D neighborhood-aware feature augmentation (3D-NAFA) module is used to augment the point features after re-projection from 2D to 3D. More specifically, a coarse-to-fine 3D voxelization strategy is leveraged to provide 3D neighborhood context to every point. Our experiments have shown that this point refinement strategy not only improves our network's segmentation performance significantly, it can be easily attached to other existing RV-based segmentation methods as a plug-and-play solution.

Our contributions can be summarized into threefold. (1) We propose a distortion-compensating knowledge distillation (DCKD) strategy to reduce the information loss due to RV projection of RGB image. (2) We devise a context-based feature fusion module to better combine the two modalities. (3) A new portable and universal point refinement scheme is proposed to fix errors in back-projection. We performed comprehensive experiments on two public datasets, SemantickKITTI \cite{semantickitti} and nuScenes \cite{nuscenes} to further validate the effectiveness of our method.

\section{Related Works}
\subsection{Camera-based Semantic Segmentation}
These methods predict the semantic labels for each pixel of the input 2D images. FCN \cite{fcn} is the pioneering work leveraging a 2D fully Convolutional Neural Network (CNN) for creating an end-to-end pipeline for image semantic segmentation. Recently, multi-scale feature learning \cite{mul1, mul2}, dilated convolution models \cite{dil1, dil2}, attention-based models \cite{att1, att2, att3} and Transformer-based models \cite{segformer, maskformer} all contributed to improvements in better segmenting 2D images. %Despite the progress, these methods do not use depth information and are sensitive to lighting condition.%, which impose challenges for AVs.

\subsection{LiDAR-based Semantic Segmentation}
These methods can be categorized based on the strategy they use to process 3D point clouds. \textbf{Point-based} methods process the unstructured 3D point clouds directly. PointNet \cite{pointnet}, uses a Multi-Layer Perception (MLP) to approximate a permutation-invariant set function. PointNet++ \cite{pointnet++} extends PointNet by sampling at different scales for extraction of relevant features both locally and globally. These methods require large memory and computational resources, especially for processing outdoor scenes \cite{tornadonet}. \textbf{Voxel-based} methods divide the 3D scene into volumetric grids known as voxels \cite{spvnas,sscn}. In order to address the varying point density and sparsity, Cylindere3D \cite{cylinder3d} uses cylindrical voxel partitions and applies asymmetrical 3D CNN to process the voxels. However, voxel-based methods are computationally expensive and their performance drops sharply when the voxel grid resolution decreases \cite{rpvnet}. \textbf{Projection-based} methods project the 3D point clouds onto 2D planes through either RV \cite{salsanet, salsanext, rangenet++}, Bird's-Eye-View (BEV) projections \cite{bev, polarnet} or both \cite{amvnet}. As projection-based methods produce compact data representation that can be processed by 2D CNN, they are more efficient and appropriate for real-time applications than point- and voxel-based methods. 

\begin{figure*}[!t]
  \centering
  \includegraphics[width=0.95\linewidth, height=6.2cm]{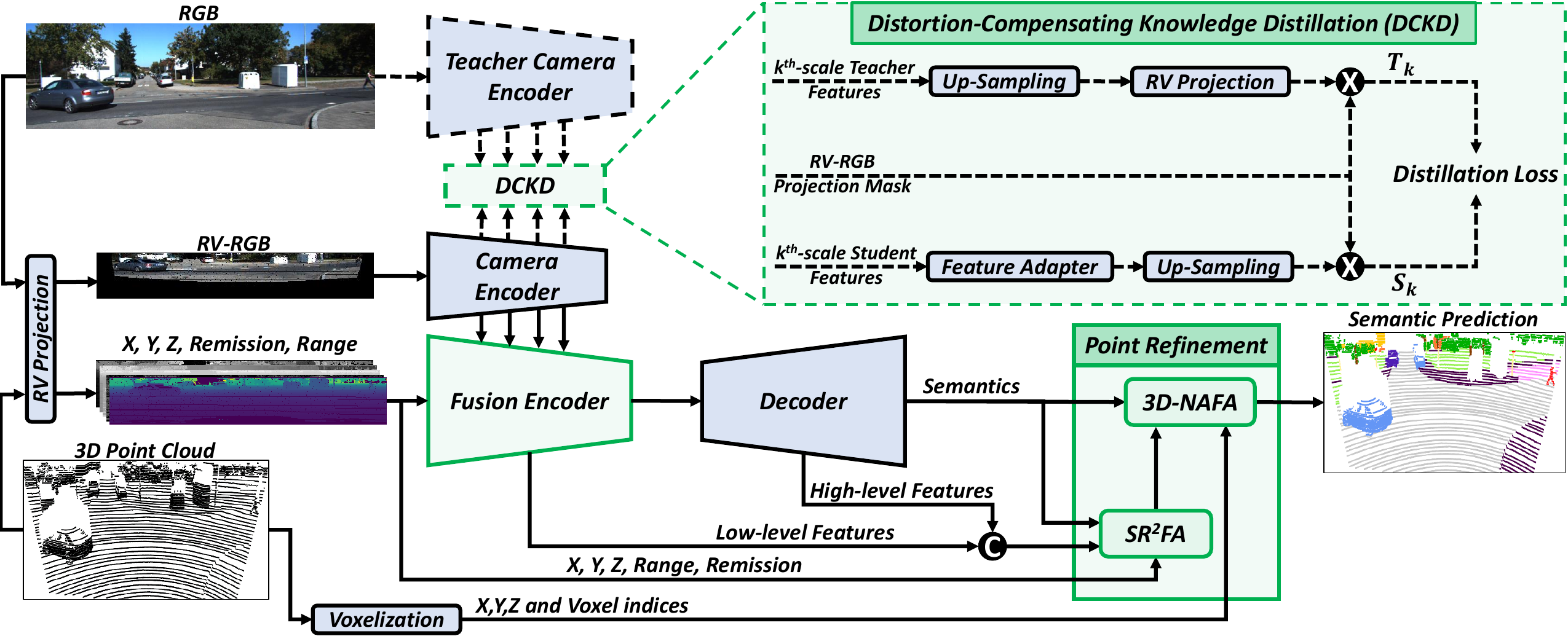}
   \caption{Overview of the proposed LaCRange framework. The blocks and arrows with dashes are only used during the training process. The blocks shown in \textcolor{DGreen}{green} are the proposed components. Best viewed in color}
   \label{fig:figure2}
   \vspace{-5mm}
\end{figure*}

\subsection{Multi-sensor Semantic Segmentation}
Multi-sensor methods attempt to properly fuse the information from both camera and LiDAR for improving accuracy and robustness of semantic segmentation \cite{fuseseg}. RGBAL \cite{rgbal} uses RV projection for both camera and LiDAR sensors and aggregates them by doing early- and mid-fusion of image and point cloud features. PMF \cite{pmf} projects the LiDAR point cloud to the camera coordinate using perspective projection and uses two separate streams for processing camera and LiDAR data by fusing the multi-sensor features in every encoder stage. However, this method is computationally expensive due to higher resolution of camera images, and the requirement to process multiple cameras. %to cover a wider Field-Of-View (FOV). 

% \subsection{KD for Semantic Segmentation}
% Knowledge distillation (KD) is one of the popular strategies to achieve model compression. In this paradigm, a large deep network (\ie teacher model) distills the knowledge into a small network (\ie student model) \cite{kd, kd2}. Many works in the image domain have leveraged KD to bridge the performance gap between the student and the teacher for semantic segmentation \cite{kd3,kd4,kd5,kd6,kd7,kd8}. To further demonstrate its versatility, \cite{pvkd} tailors KD for LiDAR semantic segmentation through both point-level and voxel-level distillation for a coarse-to-fine learning process. Extending beyond model compression, \cite{2dpass} applies knowledge distillation using 2D priors to facilitate unidirectional transfer of modal-specific knowledge in multi-sensor segmentation. % Different from the aforementioned methods, we aim to compensate the knowledge acquired by the student model that receives distorted RV-RGB images by distilling from the teacher model original distortion-free RGB images.
% % KD citations?

\section{Method}
The overview of the proposed LaCRange framework is shown in Figure \ref{fig:figure2}. Given an input LiDAR point cloud, we project it into RV image \cite{rangenet++, salsanext}, where the features are Cartesian coordinates ($x$, $y$, $z$), remission and range. Also, its corresponding RGB image(s) is projected to LiDAR coordinates using RV projection. % ONE WORD!!!

The original RGB image is passed to a pre-trained teacher encoder, while the RV-RGB image is processed by the student encoder. Extracted multi-scale teacher features are used to guide the student encoder via distortion-compensating knowledge distillation (DCKD) only during the training phase. Then, features from the student encoder are fused with the LiDAR features in the fusion encoder by context-based feature fusion modules. Finally, the RV segmentation output is processed by our proposed learnable point refinement.

\subsection{Distortion-Compensating Knowledge Distillation (DCKD)}
RV projection is suitable for real-time semantic segmentation of LiDAR point clouds due to its compact representation, but RV projection of RGB image suffers from the distortion of appearance information due to the sparsity of point clouds. We propose DCKD for alleviating this distortion in the feature space by transferring the knowledge from a teacher encoder that is pre-trained on original (distortion-free) RGB images to a student encoder that receives the corresponding RV-RGB image.

In this approach, we first pre-train a complex teacher model on a large image semantic segmentation dataset. When training the student with RV-RGB image, the frozen teacher encoder receives the corresponding original RGB image. For transferring knowledge, the student features at each encoder block need to be aligned with the teacher features. As shown in Figure \ref{fig:figure2}, this process is done in the DCKD block.

Given the  $k$th scale teacher feature maps, it is first matched to the original image resolution using bilinear up-sampling and projected to RV. Our student encoder adopts a shallower version of the teacher's architecture, hence the need for a feature adapter to match the depth with that of the teacher. In $k$th scale student features, the feature adapter block (1x1 Conv + Leaky ReLU + BatchNorm) changes the number of channels to be the same as the $k$th scale teacher, and up-samples to the original RV resolution. Both features are masked by the RV-RGB projection mask to ignore the not projected locations and obtain teacher and student features $T_{k}$ and $S_{k}$ needed for distillation. The distillation loss $L_{d}$ is obtained by calculating the average distance between $T$ and $S$ in all scales as:
\vspace{-3mm}
\begin{equation}
  L_{d} = \frac{1}{K}\sum_{k=1}^{K}||{T_{k} - S_{k}}||
  \label{eq:eq1}
  \vspace{-2mm}
\end{equation}
where $K$ is the total number of scales considered. By minimizing the above loss term, the student encoder can learn to mimic the teacher encoder. Despite receiving a distorted RGB image, the student can retrieve more coherent textural and appearance information from the teacher. In contrast to conventional KD approaches, our teacher is not only superior to the student in terms of model complexity, it is also more comprehensive in terms of the appearance integrity of its input and features.

\begin{figure}[ht]
  \centering
  \includegraphics[width=1\linewidth]{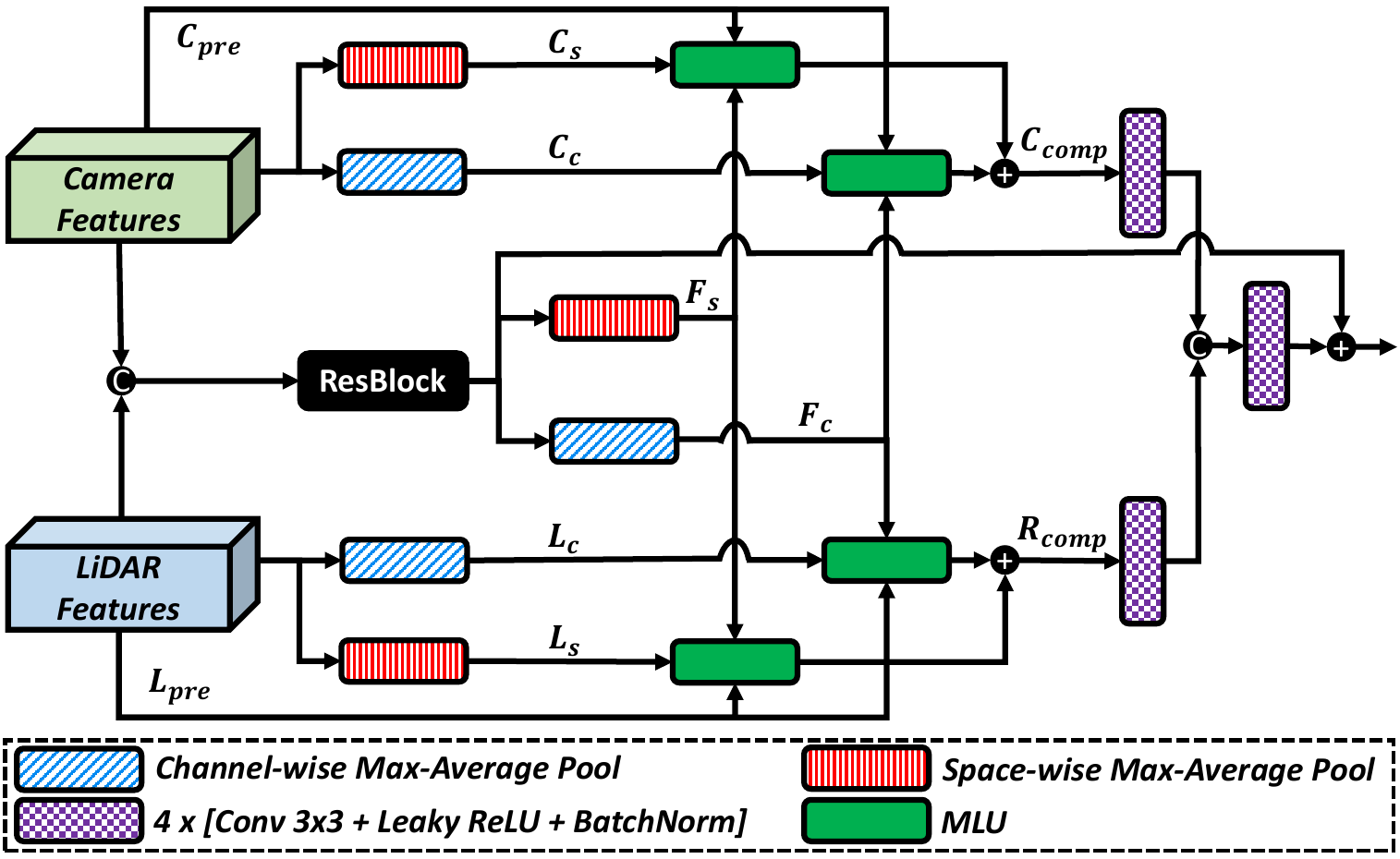}
   \caption{Context-based feature fusion module. $C_{pre}/L_{pre}$ and $C_{comp}/L_{comp}$ represent the original camera/LiDAR and retrieved camera/LiDAR features, respectively. Additionally, $C_{s}/C_{c}$, $L_{s}/L_{c}$ and $F_{s}/F_{c}$ are the space-/channel-wise pooled camera, LiDAR and initially fused features.}
   \label{fig:figure3}
   \vspace{-2mm}
\end{figure}

\subsection{Context-based Feature Fusion (CFF)}
Camera and LiDAR are two complementary sensors that each capture sensor-specific information in the scene. Preservative fusion of these two sensors is of utmost importance in multi-sensor semantic segmentation.

The block diagram of the proposed feature fusion module is shown in Figure \ref{fig:figure3}, which is comprised of two stages. In the first stage, the camera and LiDAR features are concatenated and fed to the ResBlock in the encoder for initial feature fusion. In the second stage, the initially fused features are complemented with sensor-specific information using global context of each feature, which are obtained by applying average and max pooling space- and channel-wise. The global context of initially fused features (denoted as $F_{s}/F_{c}$) can be used to interact with both camera global context (denoted as $C_{s}/C_{c}$) and LiDAR global context (denoted as $L_{s}/L_{c}$) in the Modality Look-Up block as shown in Figure \ref{fig:figure3} to retrieve each sensor's unique pre-fusion features ($C_{pre}/L_{pre}$). The captured information from each of the modalities are processed separately and fused together using several convolutional blocks and added to the initially fused features. Here, we discuss two strategies that can be used for the implementation of the MLU block, which are Convolutional Block Attention-based (CBAM) \cite{cbam} and Cross-Attention-based \cite{deepfusion}.

\subsubsection{CBAM-based Modality Look-Up}
In this strategy, the global context information of initially-fused features along space/channel ($F_{s}/F_{c}$) are added to that of the camera and LiDAR global context information ($C_{s}/C_{c}$ and $L_{s}/L_{c}$). Then, a Conv/MLP layer with a sigmoid activation is used for selecting complementary information $M_{comp}$ from the original camera and LiDAR features ($C_{pre}/L_{pre}$) as:
\vspace{-2mm}
\begin{equation}
\begin{gathered}
  M_{comp} = \Bigr[BC_{s}\left(\sigma\left(MLP\left(M_{c} + F_{c}\right)\right)\right) + \\ BC_{c}\left(\sigma\left(Conv\left(M_{s} + F_{s}\right)\right)\right)\Bigr] \times M_{pre}, \ \ M \in \{C, L\}
  \label{eq:eq9}
  \vspace{-2mm}
\end{gathered}
\end{equation}
where $M$ and $\sigma$ indicate the modality type and sigmoid activation function, respectively. $BC$ indicates the broadcasting operation along space ($BC_{s}$) and channel ($BC_{c}$).

\subsubsection{Cross-Attention-based Modality Look-Up}
In this strategy, a similar approach to \cite{deepfusion} is used. Conversely, the attention operates on the global level (spatial/channel pooled). Furthermore, we decorate the fused features with both modalities, rather than just complementing LiDAR features with camera features, as we recognize the inherent flaws of each individually. The initially fused features attend to each of the two modalities to select complementary information $M_{comp}$ as follows:
\vspace{-2mm}
\begin{equation}
\begin{gathered}
  M_{comp}^{c} = BC_{s}\left(\sigma\left(MLP\left(Att\left(Q^{c}, K_{M}^{c}, V_{M}^{c}\right)\right)\right)\right), \\
  M_{comp}^{s} = BC_{c}\left(\sigma\left(Conv\left(Att\left(Q^{s}, K_{M}^{s}, V_{M}^{s}\right)\right)\right)\right), \\
  M_{comp} = \Bigr[M_{comp}^{c} +  M_{comp}^{s}\Bigr] \times M_{pre}, \ \ M \in \{C, L\}
  \label{eq:eq3}
\end{gathered}
\end{equation}
where $Q^{c}/Q^{s}$ represent the embedded query obtained from the pooled initially fused features along channel/space, $K_{M}^{c}/K_{M}^{s}$ and $V_{M}^{c}/V_{M}^{s}$ are the key and value for the pooled features from modality $M$, respectively. $Att$ indicates the cross-attention followed by a linear layer. $M_{comp}^{c}/M_{comp}^{s}$ are the retrieved features from modality $M$ along channel/space. In the experimental results section, the effect of using each of these MLU blocks will be investigated. 

\begin{figure}[ht]
  \centering
  \includegraphics[width=1\linewidth]{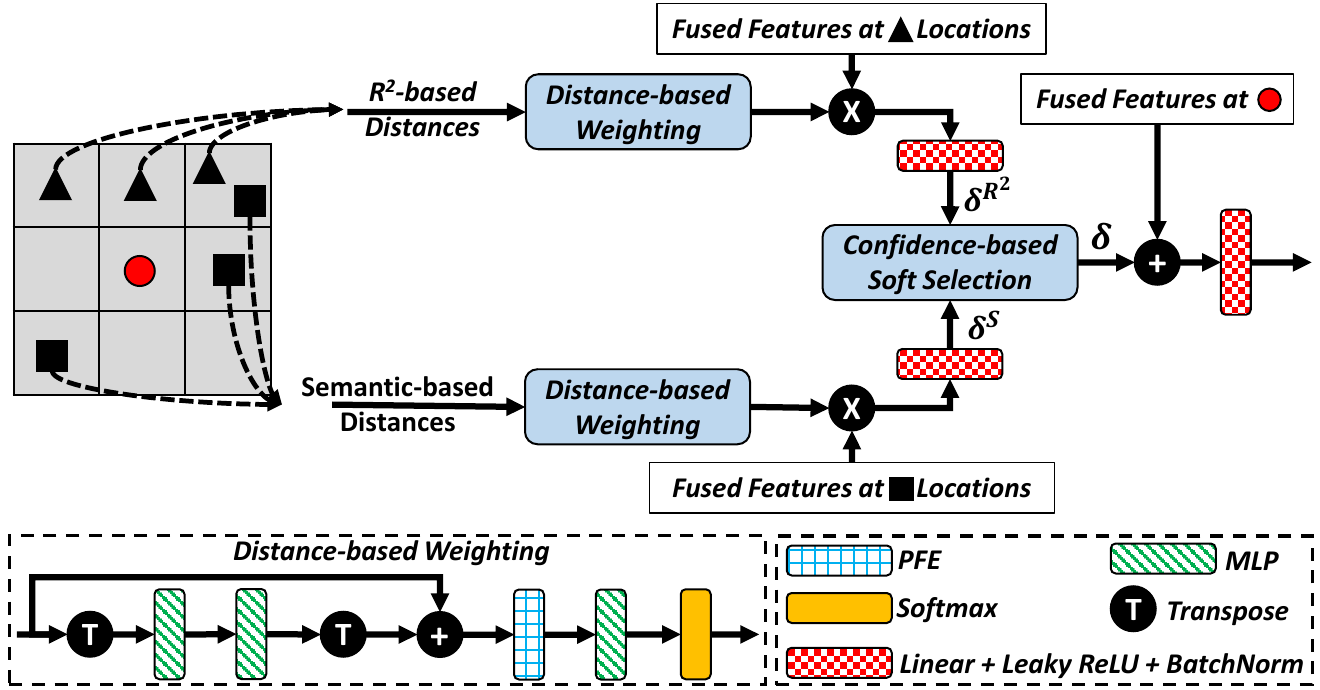}
   \caption{Semantic-Range-Remission-based Feature Aggregation (SR$^{2}$FA) module. $\blacksquare$ and $\bigblacktriangleup$ represent the locations selected based on semantic and range-remission, respectively.}
   \label{fig:figure4}
   \vspace{-3mm}
\end{figure}

\subsection{Point Refinement}
Despite real-time performance advantage of RV-based segmentation methods \cite{rangenet++, salsanext}, their accuracy is limited. Motivated by this limitation, a new point refinement scheme is proposed, which includes a feature aggregation on the RV-level (SR$^{2}$FA) and a feature augmentation on the point-level (3D-NAFA).
%which acts as a plug-and-play solution for boosting the segmentation accuracy of any RV-based segmentation method.

\subsubsection{Semantic-Range-Remission-based Feature Aggregation (SR$^{2}$FA)}
Due to down-sampling operations in the encoder, some useful low-level geometry information is lost. To retrieve this information, we fuse low-level features before entering the encoder with the features from the last block of the decoder. These features need to be aggregated locally based on the context using operations such as convolution, but the restrictions in geometric transformation modeling capacity of the standard convolution due to fixed grid structure makes it a naive approach \cite{efficientlps}.

Here, we designed the SR$^{2}$FA module, which is shown in Figure \ref{fig:figure4}. The RV projection mask is used to filter out the un-projected locations on the RV plane. Then, for each target pixel, the nearest neighboring pixels are selected twice based on different selection metrics: estimated semantic probability absolute difference (S) and range-remission absolute difference (R$^{2}$). For the neighbors selected based on range-remission, their absolute differences in terms of xyz, range, remission and RGB with respect to the center are used to form a set of feature distances. For the neighbors selected based on semantic probability, their semantic probability absolute differences are used to form another set of feature distances. These feature distances are passed through a Point Feature Encoder (PFE) and MLP layers for individual processing of each neighbor. PFE is composed of an MLP layer followed by max- and average-pool layers and outputs the concatenation of processed features with the average- and max-pooled features along neighbors. Finally, a Softmax function is applied in each branch to produce a set of weights, which are used for calculating weighted sums of selected neighbors' features $\delta^{S}$ and $\delta^{R^{2}}$ that are based on semantics and range-remission, respectively.

For combining these two sets of aggregated features ($\delta^{S}$ and  $\delta^{R^{2}}$), a dynamic confidence-based soft selection procedure is designed. In this procedure, the mean semantic confidence $\Phi_{m,n}$, of the neighborhood for every center pixel is calculated as follows: 
%\begin{equation}
%\begin{gathered}
%  \Phi_{i,j} = 1 + \left[\frac{1}{\log{C}} \sum_{c=1}^{C} S_{i,j,c} \log{S_{i,j,c}}\right], \\
%  \overline{\Phi}_{i,j} = \frac{1}{N_{\psi}} \sum_{m,n \in \psi(i,j)} \Phi_{m,n} 
%  \label{eq:eq5}
%\end{gathered}
%\end{equation}
\vspace{-2mm}
\begin{equation}
\begin{gathered}
  \Phi_{m,n} = \frac{1}{N_{\psi(m,n)}}\sum_{(i,j)\in \psi(m,n)} \left[1 + \frac{1}{\log{C}} \sum_{c=1}^{C} S_{i,j,c} \log{S_{i,j,c}}\right] \\
  %\overline{\Phi}_{i,j} = \frac{1}{N_{\psi}} \sum_{m,n \in \psi(i,j)} \Phi_{m,n} 
  \label{eq:eq5}
\end{gathered}
% \vspace{-1mm}
\end{equation}
where $S_{i,j,c}$ is the estimated semantic probability for $c$th class on location $(i,j)$. $\psi(m,n)$ and $N_{\psi(m,n)}$ indicate the neighboring coordinates of $(m,n)$ and number of neighbors, respectively. %big N, as well as notation
The two sets of aggregated features are combined to produce $\delta$, the confidence-based fused features as:
\vspace{-2mm}
\begin{equation}
\begin{gathered}
  \delta = \left[1-\exp^{-\lambda(1-\Phi)}\right]\delta^{R^{2}} + \left[\exp^{-\lambda (1-\Phi)}\right]\delta^{S} \\
  %F = \delta + F'
  \label{eq:eq6}
\end{gathered}
\vspace{-2mm}
\end{equation}
where $\lambda$ is a hyper-parameter controlling the mean confidence effect. Based on the above equation, in the neighborhoods with low mean semantic confidence, the aggregated features from the semantic branch are down-weighted while the range-remission branch aggregated features are attended more, and vice-versa. The aggregated features $\delta$ are added to the center feature at $(m,n)$ and further processed to yield the final features on RV image.
\begin{figure}[!t]
  \centering
  \includegraphics[width=1\linewidth]{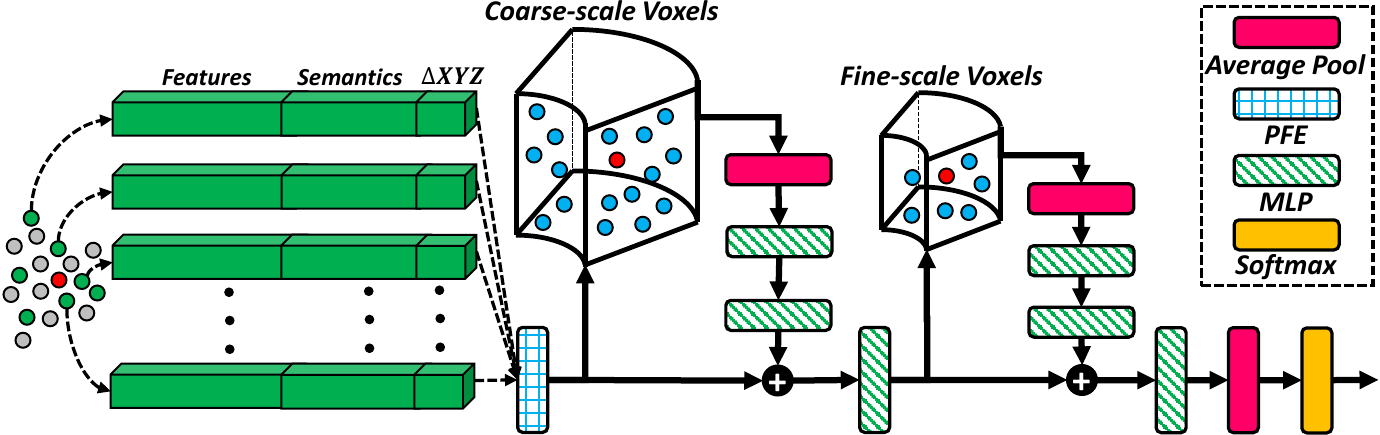}
   \caption{3D Neighborhood-Aware Feature Augmentation (3D-NAFA) module. Best viewed in color.}
   \label{fig:figure5}
   \vspace{-5mm}
\end{figure}

\subsubsection{3D Neighborhood-Aware Feature Augmentation (3D-NAFA)}
Solely re-projecting RV features to 3D neglects scene topology and relations of point geometries. To further augment the features of each point to be more 3D neighborhood-aware, a coarse-to-fine augmentation process is utilized. 

\begin{table*}[!t]
  \centering
    \caption{Results on SemanticKITTI \cite{semantickitti} validation set. ($\ast$) Results obtained from \cite{pmf}. L and C denote LiDAR and camera, respectively.}
    \vspace{-4mm}
    \resizebox{2\columnwidth}{!}{
  \begin{tabular}{ |*{24}{c|} }
    \toprule
    Method & Input & \rotatebox[origin=c]{45}{mIOU} & \rotatebox[origin=c]{45}{car} & \rotatebox[origin=c]{45}{bicycle} & \rotatebox[origin=c]{45}{motorcycle} & \rotatebox[origin=c]{45}{truck} & \rotatebox[origin=c]{45}{other-vehicle} & \rotatebox[origin=c]{45}{person} & \rotatebox[origin=c]{45}{bicyclist} & \rotatebox[origin=c]{45}{motorcyclist} & \rotatebox[origin=c]{45}{road} & \rotatebox[origin=c]{45}{parking} & \rotatebox[origin=c]{45}{sidewalk} & \rotatebox[origin=c]{45}{other-ground} & \rotatebox[origin=c]{45}{building} & \rotatebox[origin=c]{45}{fence} & \rotatebox[origin=c]{45}{vegetation} & \rotatebox[origin=c]{45}{trunk} & \rotatebox[origin=c]{45}{terrain} &  \rotatebox[origin=c]{45}{pole} & \rotatebox[origin=c]{45}{traffic-sign} & \rotatebox[origin=c]{45}{speed (ms)}\\
    \midrule
    \midrule
    %RandLANet & L & 50.0 & 92.0 & 8.0 & 12.8 & 74.8 & 46.7 & 52.3 & 46.0 & 0.0 & 93.4 & 32.7 & 73.4 & 0.1 & 84.0 & 43.5 & 83.7 & 57.3 & 73.1 & 48.0 & 27.3 & - \\
    RangeNet++ \textsuperscript{$\ast$} \cite{rangenet++} & L & 51.2 & 89.4 & 26.5 & 48.4 & 33.9 & 26.7 & 54.8 & 69.4 & 0.0 & 92.9 & 37.0 & 69.9 & 0.0 & 83.4 & 51.0 & 83.3 & 54.0 & 68.1 & 49.8 & 34.0 & 17.1 \\
    %SequeezeSegV2 & L & 40.8 & 82.7 & 15.1 & 22.7 & 25.6 & 26.9 & 22.9 & 44.5 & 0.0 & 92.7 & 39.7 & 70.7 & 0.1 & 71.6 & 37.0 & 74.6 & 35.8 & 68.1 & 21.8 & 22.2 & - \\
    SqueezeSegV3 \textsuperscript{$\ast$} \cite{squeezesegv3} & L & 53.3 & 87.1 & 34.3 & 48.6 & 47.5 & 47.1 & 58.1 & 53.8 & 0.0 & 95.3 & 43.1 & 78.2 & 0.3 & 78.9 & 53.2 & 82.3 & 55.5 & 70.4 & 46.3 & 33.2 & 40.9 \\
    PolarNet \cite{polarnet} & L & 56.4 & 93.2 & 43.7 & 46.5 & 56.3 & 30.6 & 67.2 & 78.2 & 0.0 & 94.1 & 29.3 & 75.0 & 0.1 & 87.6 & 49.9 & 86.4 & 61.7 & 72.5 & 59.2 & 40.7 & 15.5 \\        
    SalsaNext \textsuperscript{$\ast$} \cite{salsanext} & L & 59.4 & 90.5 & 44.6 & 49.6 & 86.3 & 54.6 & 74.0 & 81.4 & 0.0 & 93.4 & 40.6 & 69.1 & 0.0 & 84.6 & 53.0 & 83.6 & 64.3 & 64.2 & 54.4 & 39.8 & \textbf{8.0} \\
    %MinkowskiNet & 58.5 & 95.0 & 23.9 & 50.4 & 55.3 & 45.9 & 65.6 & 82.2 & 0.0 & 94.3 & 43.7 & 76.4 & 0.0 & 87.9 & 57.6 & 87.4 & 67.7 & 71.5 & 63.5 & 43.6 & ? \\
    %SPVNAS & 62.3 & \textbf{96.5} & 44.8 & 63.1 & 59.9 & 64.3 & 72.0 & 86.0 & 0.0 & 93.9 & 42.4 & 75.9 & 0.0 & 88.8 & 59.1 & 88.0 & 67.5 & 73.0 & 63.5 & \textbf{44.3} & ? \\
    %Cylinder3D & 64.9 & 96.4 & \textbf{61.5} & \textbf{78.2} & 66.3 & 69.8 & \textbf{80.8} & \textbf{93.3} & 0.0 & 94.9 & 41.5 & 78.0 & 1.4 & 87.5 & 50.0 & 86.7 & 72.2 & 68.8 & 63.0 & 42.1 & ? \\
    PointPainting \textsuperscript{$\ast$} \cite{pointpainting} & L + C & 54.5 & 94.7 & 17.7 & 35.0 & 28.8 & 55.0 & 59.4 & 63.6 & 0.0 & 95.3 & 39.9 & 77.6 & 0.4 & 87.5 & 55.1 & 87.7 & 67.0 & 72.9 & 61.8 & 36.5 & - \\
    RGBAL \textsuperscript{$\ast$} \cite{rgbal} & L + C & 56.2 & 87.3 & 36.1 & 26.4 & 64.6 & 54.6 & 58.1 & 72.7 & 0.0 & 95.1 & 45.6 & 77.5 & \textbf{0.8} & 78.9 & 53.4 & 84.3 & 61.7 & 72.9 & 56.1 & 41.5 & 9.8 \\
    PMF \cite{pmf} & L + C & 63.9 & \textbf{95.4} & 47.8 & \textbf{62.9} & 68.4 & \textbf{75.2} & \textbf{78.9} & 71.6 & 0.0 & \textbf{96.4} & 43.5 & \textbf{80.5} & 0.1 & 88.7 & 60.1 & \textbf{88.6} & \textbf{72.7} & \textbf{75.3} & \textbf{65.5} & \textbf{43.0} & 83.8 \\  
    \rowcolor{Gray}
    \textbf{LaCRange (Ours)} & L + C & \textbf{64.1} & 95.2 & \textbf{51.5} & 46.0 & \textbf{86.6} & 60.4 & 76.4 & \textbf{88.9} & 0.0 & 95.7 & \textbf{45.5} & 79.9 & 0.0 & \textbf{89.2} & \textbf{63.8} & 88.2 & 68.8 & 73.8 & \textbf{65.5} & 42.7 & 49.5 \\
    \bottomrule
  \end{tabular}
  }
  \label{tab:kittivalset}
  \vspace{-2mm}
\end{table*}

%PolarNet(Results&Time), RPVNet(Results&Time), AF2S3Net(Results&Time), SPVNASS (Time), Cylinder3D(Time), MinkowskiNet(Time), Other 2D methods.

\begin{table*}[!t]
  \centering
    \caption{Semantic segmentation results on the nuScenes \cite{nuscenes} test benchmark. (\textdagger) Time reported, only includes the 6 image passes to the model per LiDAR scan and the prediction merging step is not included. ($\ddagger$) We trained and submitted for evaluation.}
    \vspace{-4mm}
    \resizebox{2\columnwidth}{!}{
  \begin{tabular}{ |*{22}{c|} }
    \toprule
    Method & Input & \rotatebox[origin=c]{45}{mIOU} & \rotatebox[origin=c]{45}{FW mIOU} & \rotatebox[origin=c]{45}{barrier} & \rotatebox[origin=c]{45}{bicycle} & \rotatebox[origin=c]{45}{bus} & \rotatebox[origin=c]{45}{car} & \rotatebox[origin=c]{45}{construction} & \rotatebox[origin=c]{45}{motorcycle} & \rotatebox[origin=c]{45}{pedestrian} & \rotatebox[origin=c]{45}{traffic cone} & \rotatebox[origin=c]{45}{trailer} & \rotatebox[origin=c]{45}{truck} & \rotatebox[origin=c]{45}{drivable} & \rotatebox[origin=c]{45}{Other-flat} & \rotatebox[origin=c]{45}{side walk} & \rotatebox[origin=c]{45}{terrain} & \rotatebox[origin=c]{45}{manmade} & \rotatebox[origin=c]{45}{vegetation} &
    \rotatebox[origin=c]{45}{speed (ms)} \\
    \midrule
    \midrule
    SalsaNext \textsuperscript{$\ddagger$} \cite{salsanext} & L & 66.8 & 85.0 & 68.9 & 13.8 & 78.0 & 78.2 & 43.1 & 64.4 & 59.3 & 58.3 & 74.9 & 59.8 & 95.1 & 63.7 & 73.4 & 69.8 & 85.0 & 82.3 & \textbf{8.4} \\
    PolarNet \cite{polarnet} & L & 69.4 & 87.4 & 72.2 & 16.8 & 77.0 & \textbf{86.5} & 51.1 & 69.7 & 64.8 & 54.1 & 69.7 & 63.5 & 96.6 & 67.1 & 77.7 & 72.1 & 87.1 & 84.5 & 32.1 \\
    %LiFusion & L + C & 75.7 & 58.1 & 36.3 & 86.7 & 84.3 & 60.0 & 79.7 & 80.3 & 77.8 & 83.2 & 68.7 & 97.2 & 68.2 & 77.0 & 74.5 & 91.0 & 89.0 & - \\
    PMF-ResNet34 \cite{pmf} & L + C & 75.5 & 88.9 & 80.2 & 35.7 & 79.7 & 86.0 & 62.5 & 76.4 & 77.0 & 73.7 & 78.5 & 66.9 & 97.1 & 65.3 & 77.6 & 74.4 & 89.5 & 87.7 & 1505.7\textsuperscript{\textdagger} \\ 
    PMF-ResNet50 \cite{pmf}& L + C & \textbf{77.0} & \textbf{89.4} & \textbf{82.1} & \textbf{40.3} & 80.9 & 86.4 & 63.7 & 79.2 & \textbf{79.8} & \textbf{75.9} & \textbf{81.2} & 67.1 & \textbf{97.3} & \textbf{67.7} & 78.1 & 74.5 & 90.0 & \textbf{88.5} & 1883.3\textsuperscript{\textdagger} \\ 
    \rowcolor{Gray}
    \textbf{LaCRange (Ours)} & L + C & 75.3 & 89.1 & 78.0 & 32.6 & \textbf{88.3} & 84.5 & \textbf{63.9} & \textbf{81.5} & 75.6 & 72.5 & 64.7 & \textbf{68.0} & 96.6 & 65.9 & \textbf{78.6} & \textbf{75.0} & \textbf{90.4} & 88.3 & 50.1 \\
    \bottomrule
  \end{tabular}
  }
  \label{tab:nuscenestest}
  \vspace{-3mm}
\end{table*}

The diagram of the 3D-NAFA is shown in Figure \ref{fig:figure5}. For every 3D point and its projection location on the RV image, its $K$ nearest neighbors in terms of range are considered. For every selected neighbor, its features obtained from SR$^{2}$FA module, estimated semantics, and relative 3D distance from the target point are concatenated and passed to a PFE layer.
%\vspace{-3mm}
%\begin{equation}
%\begin{gathered}
%  {P'_{i}}^{k} =  PFE\left({P_{i}}^{k}\right)
%  \label{eq:eq8}
%\end{gathered}
%\vspace{-1mm}
%\end{equation}
%where ${P_{i}}^{k}$ is the concatenated features. 

These features only consider the 2D neighborhood for assigning a feature vector to a point, yet the selected neighbors are not necessarily the nearest in 3D.
Therefore, we augment the point features by injecting multi-scale 3D neighborhood information. Here, a multi-scale coarse-to-fine cylindrical voxelization \cite{cylinder3d} is leveraged for 3D point feature augmentation.
Specifically, the 3D scene is first voxelized with a coarse-scale voxel grid, and all the points in each voxel are average-pooled and passed through two MLP layers, which are then broadcast to every point inside the voxel to generate the coarsely aggregated features.
% \vspace{-2mm}
% \begin{equation}
% \begin{gathered}
%   {P_{i, s_{c}}} = BC\left(MLP\left(MLP\left(\frac{1}{N_{j}} \sum_{i=1}^{N_{j}} {P_{i}}\right)\right)\right)%, \\
%   %P_{{V_{j}}^{s_{c}}} = Broadcast\left({V_{j}}^{s_{c}}\right)
%   \label{eq:eq9}
% \end{gathered}
% \vspace{-2mm}
% \end{equation}
The point features are added to these aggregated features with a skip connection and passed through a MLP layer to generate coarsely augmented features.
% \vspace{-2mm}
% \begin{equation}
% \begin{gathered}
%   \widehat{P}_{i, s_{c}} = MLP\left({P_{i}} + P_{i,s_{c}}\right) 
%   \label{eq:eq10}
% \end{gathered}
% \vspace{-3mm}
% \end{equation}
As shown in Figure \ref{fig:figure5}, the same process is repeated but with a fine-scale voxel grid for augmenting the coarsely augmented point features with a more fine-grained 3D neighborhood. Finally, the multi-scale augmented point features are average-pooled along the $K$ neighbors and passed through a softmax function to provide the point labels. It will be shown in the experimental results section that overall point refinement module can be attached to any RV-based semantic segmentation method as a plug-and-play solution to uplift RV predictions.

\section{Experiments}
\subsection{Dataset and Metric}
\textbf{SemanticKITTI \cite{semantickitti}} is a large-scale dataset for dense semantic segmentation task that is captured by 64 beams LiDAR sensor. Although the LiDAR sensor provides $360^{\circ}$ FOV, it only includes the front-view camera limited to $90^{\circ}$.

\textbf{nuScenes \cite{nuscenes}} is another popular large-scale autonomous driving dataset that includes 1,000 driving scenes captured in different cities and under different illumination and weather conditions. The data is recorded by a 32 beams LiDAR sensor and a total of six cameras that cover the full $360^{\circ}$. 

Similar to other segmentation methods, the mean intersection over union (mIoU) is used as evaluation metric, which is defined as the average IoU over all classes.

\subsection{Implementation Details}
For the implementation, we used the Pytorch deep learning library \cite{pytorch}. For the teacher-student image encoders, we relied on ResNet family architectures \cite{resnet}. For any teacher encoder architecture, the respective student's model number of channels are halved. In all experiments, the teacher and student encoders are based on ResNet18 unless otherwise specified. Under nuScenes- and SemanticKITTI-based experiments, the teacher model is pre-trained on nuImages dataset \cite{nuscenes}, and KITTI-STEP dataset \cite{kittistep}, respectively.

\subsection{Results on SemanticKITTI}
Similar to PMF \cite{pmf}, as SemanticKITTI \cite{semantickitti} only provides the images of the front-view camera, we only evaluate on the colored points. This is obtained by projection of point clouds to the camera coordinates and only keeping the points that have corresponding RGB values. We compare LaCRange with several state-of-the-art single-view 2D methods. These results are shown in Table \ref{tab:kittivalset}. As can be observed, our performance surpasses all compared methods in terms of mIOU. Compared to all RV-based methods, LaCRange obtains higher IOU in majority of the classes. It is also worth noting that LaCRange performs much better or on par with PMF \cite{pmf} (which does not rely on RV projection) in some small object classes with complex shapes such as bicycle, bicyclist, pole, and traffic-sign. These classes are more sensitive to distortion in RV projection and errors in kNN when re-projecting to 3D.

\subsection{Results on nuScenes}
The results of our method and state-of-the-art published single-view 2D methods on nuScenes test set are shown in Table \ref{tab:nuscenestest}. These results demonstrate that LaCRange has significantly higher mIOU than the LiDAR-only models. Furthermore, LaCRange performs on par with PMF-ResNet34 but worse than PMF-ResNet50, yet it should be noted that it runs considerably faster. This is mainly because PMF \cite{pmf} needs to process 6 camera images for each LiDAR frame to fully cover $360^{\circ}$ FOV, while our method only processes a single RV image for the same FOV. 

\subsection{Ablation Studies}

\begin{table}[t]
  \centering
  \caption{Ablation study on the SemanticKITTI \cite{semantickitti} validation set.}
    \vspace{-3mm}
  \begin{tabular}{*{6}{c}}
    \toprule
    Method & DCKD & CFF & 3D-NAFA & SR$^{2}$FA & mIOU \\
    \midrule
    \midrule
    \rowcolor{Gray}
    Baseline &  &  &  &  & 57.8  \\
     & \checkmark  &  &  &  & 58.5 \\
     & \checkmark & \checkmark &  &  & 59.5  \\
     & \checkmark & \checkmark & \checkmark &  & 64.0 \\
     & \checkmark & \checkmark & \checkmark & \checkmark & \textbf{64.1} \\
    \bottomrule
  \end{tabular}
  \label{tab:components}
  \vspace{-7mm}
\end{table}
\subsubsection{Effects of Network Components}
As shown in Table \ref{tab:components}, each of our new components amounts to an improvement in the overall mIoU. The biggest gain is obtained by the 3D-NAFA module leading to $4.5\%$ increase. The context-based feature fusion module introduced $1.0\%$ higher mIOU. Moreover, the DCKD strategy brings $0.7\%$ mIoU improvement. Note that although the SR$^{2}$FA module only shows $ 0.1\%$ improvement in mIOU, it brings $0.6\%$ increase in mIOU on nuScenes validation set. This is because nuScenes \cite{nuscenes} was collected using 32-beam LiDAR rather than the 64-beam LiDAR of SemanticKITTI \cite{semantickitti}, resulting in much sparser point clouds that benefit more from neighborhood information aggregation.

\begin{table}[!h]
  \vspace{-3mm}
  \centering
  \caption{Different teacher-student architectures in DCKD.}
  \vspace{-3mm}
  \begin{tabular}{*{3}{c}}
    \toprule
    Method & Encoder & mIOU \\
    \midrule
    \midrule
    \rowcolor{Gray}
    Student & ResNet18-Like & 57.8  \\
    Student(DCKD) & ResNet18-Like & \textbf{58.5} (\textbf{+0.7})  \\
    \midrule
    \rowcolor{Gray}
    Student & ResNet34-Like & 57.8  \\
    Student(DCKD) & ResNet34-Like & \textbf{58.7} (\textbf{+0.9})  \\
    %\midrule
    %\rowcolor{Gray}
    %Student & ResNet50-Like & 57.8  \\
    %Student(DCKD) & ResNet50-Like & \textbf{57.9} (\textbf{+0.1})  \\    
    \bottomrule
  \end{tabular}
  \label{tab:studentteacher}
  \vspace{-3mm}
\end{table}

% \begin{figure}[!t]
%   \centering
%   \includegraphics[width=1\linewidth]{DCKD.pdf}
%    \caption{Performances of different image encoders in terms of mIOU and time on nuScenes \cite{nuscenes} validation set.}
%    \label{fig:figure6}
%    \vspace{-4mm}
% \end{figure}

% \begin{figure}[!t]
%   \centering
%   \includegraphics[width=1\linewidth, height=4cm]{DCKD_Similarity.pdf}
%    \caption{Similarity map calculated between student and teacher feature maps. From top to bottom: Original RGB, RV-RGB, Similarity map. Best viewed in color.}
%    \label{fig:figure7}
%    \vspace{-3mm}
% \end{figure}

\subsubsection{Effects and Analyses of DCKD}
In Table \ref{tab:studentteacher}, the effects of using different teacher-student architectures are shown. It can be seen that the DCKD is able to consistently improve the performance of the student network. It should also be noted that DCKD does not introduce any additional latency compared to the standalone student.
% In Figure \ref{fig:figure6}, the results of using different networks for image encoder are shown on nuScenes dataset. It can be seen that the pre-trained teacher outperforms student trained from scratch as well as student trained with DCKD, while its inference time is significantly higher. On the other hand, student trained with DCKD shows 1.3\% improvement in mIOU compared to student trained without teacher guidance while having the same inference time.

% In Figure \ref{fig:figure7}, the similarity map between the teacher and student features is shown, where a distorted bicyclist (red-colored) patch from the student feature maps is compared with the RV-projected teacher feature maps via a sliding window.
% The highest similarity is obtained on the same location as the bicyclist, illustrating that the student is able to mimic the teacher. Moreover, a high correlation between the two bicyclist (red- \& green-colored) patches can be seen, which implies the semantic information distilled from the teacher model to the student is well-received\footnote{More results are shown in supplementary materials.}.
\begin{table}[!h]
  \vspace{-2mm}
  \centering
  \caption{Different MLU strategies in camera-LiDAR feature fusion.}
  \vspace{-3mm}
  \resizebox{0.8\columnwidth}{!}{
  \begin{tabular}{*{3}{c}}
    \toprule
    Method & mIOU $\uparrow$ & Inference Time $\downarrow$ \\
    \midrule
    \midrule
    \rowcolor{Gray}
    Baseline & 58.5 & 23.1 ms  \\
    CBAM-based Look-Up & 59.2 (+0.7) & 29.1 ms \\
    Cross-Attention-based Look-Up & 59.0 (+0.5) & 45.5 ms  \\
    Combined & \textbf{59.5} (\textbf{+1.0}) & 31.1 ms  \\
    \bottomrule
  \end{tabular}
  }
  \label{tab:featurefusion}
  \vspace{-3mm}
\end{table}
\subsubsection{Modality Look-Up Strategy}
The results of using different MLU strategies in the fusion module is shown in Table \ref{tab:featurefusion}. Both CBAM-based as well as cross-attention-based strategies show mIOU gain over the baseline. However, the computational complexity and memory usage of cross-attention calculation increases with feature resolution. By combining these two approaches (2 CBAM + 2 Cross Attention),
%By combining these two approaches (first two fusion module based on CBAM and the later two based on cross-attention),
not only is the model more mobile due to less memory usage, the inference time is also reduced. The mIOU improvement is more significant because higher-level features encapsulate more global context, which can be captured by the global receptive field of the cross-attention. 
\begin{table}[!t]
  \centering
  \caption{Effects of different scales and sizes of voxel grids.}
  \vspace{-3mm}
  \resizebox{0.7\columnwidth}{!}{
  \begin{tabular}{*{3}{c}}
    \toprule
    Grid Size(s) & mIOU $\uparrow$ & Inference Time $\downarrow$ \\
    \midrule
    \midrule
    $[480, 90, 64]$ & 63.2 & 40.4 ms  \\
    $[30, 3, 2]$ & 63.5 & 39.7 ms  \\
    \midrule
    $[360, 45, 32], [480, 90, 64]$ & 63.4 & 44.9 ms  \\
    % $[240, 23, 16], [360, 45, 32]$ & 63.7 & 44.3 ms  \\
    $[120, 12, 8], [240, 23, 16]$ & 63.9 & 44.0 ms  \\
    % $[60, 6, 4], [120, 12, 8]$ & 63.9 & 43.9 ms  \\
    $[30, 3, 2], [60, 6, 4]$ & \textbf{64.0} & 43.8 ms  \\
    %$[30, 3, 2], [120, 12, 8]$ & 63.9 & 43.4 ms  \\
    \bottomrule
  \end{tabular}
  }
  \label{tab:3dnafa}
  \vspace{-3mm}
\end{table}

\subsubsection{Point Refinement Design Analysis}
\begin{figure}[!h]
  \vspace{-1mm}
  \centering
  \includegraphics[width=1\linewidth, height=3.5cm]{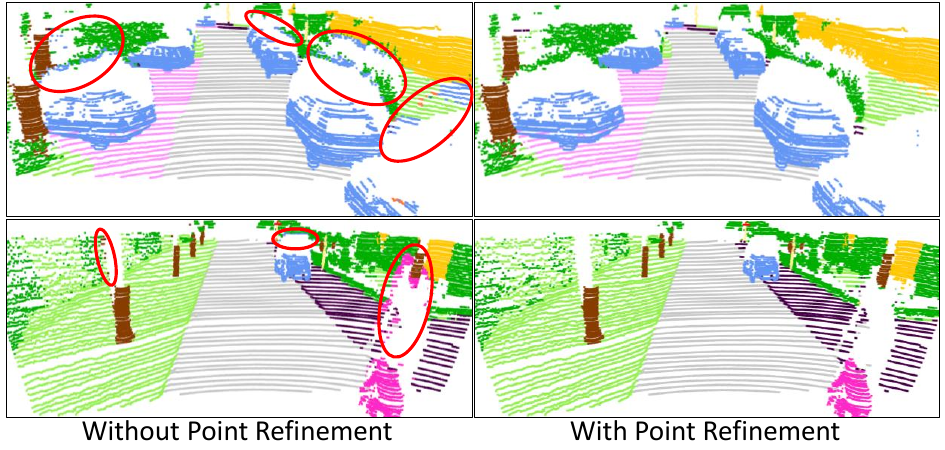}
   \vspace{-8mm}
   \caption{Results of segmentation with (right) and without proposed point refinement (left). Best viewed in color.}
   \label{fig:figure8}
\end{figure}

%\subsubsection{Effects of Various Voxel Sizes and Scales}
We investigated the effect of different voxel grid scales and sizes in 3D-NAFA performance as shown in Table \ref{tab:3dnafa}. In all experiments, coarser grid sizes result in better mIOU due to larger neighborhoods that provide interactive information to correct wrong labels. Moreover, multi-scale voxel grid achieves a better mIOU than the single-scale grid with only slight inference time increase. In Figure \ref{fig:figure8}, the effects of the proposed point refinement module are shown. As can be seen, it is able to correct labels near the border of objects after re-projection from 2D to 3D.
% \begin{table}
%   \centering
%   \caption{Effects of semantic and range-remission in SR$^{2}$FA for performance improvement.}
%   \begin{tabular}{*{3}{c}}
%     \toprule
%     Semantic & Range-Remission & mIOU \\
%     \midrule
%     \midrule
%     \checkmark & \xmark & 63.9 \\
%     \xmark & \checkmark & 63.5 \\
%     \checkmark & \checkmark & \textbf{64.1} \\
%     \bottomrule
%   \end{tabular}
%   \label{tab:sr2fa}
%   \vspace{-1mm}
% \end{table}
%\subsubsection{Semantic vs. Range-Remission in SR$^{2}$FA}
% In Table \ref{tab:sr2fa}, the effectiveness of neighbor selection strategies are shown. It can be noticed that the combination of both Semantic- and Range-Remission-based neighbor selection strategies results in better segmentation performance. 

\vspace{-1mm}

\begin{table}[!t]
  \centering
  \caption{Effects of using the proposed point refinement block in other RV-based LiDAR semantic segmentation methods.}
  \vspace{-4mm}
  \resizebox{0.7\columnwidth}{!}{
  \begin{tabular}{*{3}{c}}
    \toprule
    Method & Point Refinement & mIOU \\
    \midrule
    \midrule
    % \multirow{2}{*}{RangeNet++ \cite{rangenet++}} & \xmark & 47.1 \\
    %  & \checkmark & \textbf{52.0} (\textbf{+4.9}) \\
    %  \midrule
    \multirow{2}{*}{SqueezeSegV3 \cite{squeezesegv3}} & \xmark & 48.1 \\
     & \checkmark & \textbf{52.5} (\textbf{+4.4}) \\     
     \midrule     
    \multirow{2}{*}{SalsaNext \cite{salsanext}} & \xmark & 55.8 \\
     & \checkmark & \textbf{61.1} (\textbf{+5.3}) \\
    \bottomrule
  \end{tabular}
  }
  \label{tab:pr}
  \vspace{-6mm}
\end{table}

\subsubsection{Universality of Point Refinement}
%\subsubsection{Point Refinement on other RV-based Semantic Segmentation}
We performed a set of experiments to investigate the effectiveness of the proposed point refinement scheme on improving the accuracy of the other well-known RV-based semantic segmentation methods, namely SqueezeSegV3 \cite{squeezesegv3} and SalsaNext \cite{salsanext}. We fine-tuned our point refinement module on these methods for 60 epochs. Table \ref{tab:pr} compares results of the proposed point refinement module attached to different methods. The results demonstrate that the effectiveness of this module is universal across different RV-based segmentation methods.

\section{Conclusion}
In this paper, we have proposed LaCRange, a LiDAR and Camera Range-view-based point cloud semantic segmentation method. Specifically, for alleviating the adverse effect of RV projection of RGB images, a distortion-compensating knowledge distillation (DCKD) strategy is proposed. Moreover, a context-based feature fusion module is proposed to preserve the sensor-specific information through Modality Look-Up. Finally, a new portable and universal point refinement scheme is devised to address the shortcomings of RV projection. Extensive experiments are performed on SemanticKITTI and nuScenes datasets to demonstrate the effectiveness of each proposed component.


\begin{thebibliography}{99}

\bibitem{ade20k} B. Zhou, H. Zhao, X. Puig, S. Fidler, A. Barriuso, A. Torralba, Scene parsing through ade20k dataset, In Proceedings of the IEEE/CVF Conference on Computer Vision and Pattern Recognition, pp. 633-641, 2017. 

\bibitem{cityscapes} M. Cordts, M. Omran, S. Ramos, T. Rehfeld, M. Enzweiler, R. Benenson, U. Franke, Uwe S. Roth, B. Schiele, The cityscapes dataset for semantic urban scene understanding, In Proceedings of the IEEE/CVF Conference on Computer Vision and Pattern Recognition, pp. 3213-3223, 2016.	

\bibitem{coco-stuff} H. Caesar, J. Uijlings, V. Ferrari, Coco-stuff: Thing and stuff classes in context, In Proceedings of the IEEE/CVF Conference on Computer Vision and Pattern Recognition, pp. 1209-1218, 2018.

\bibitem{nuscenes} H. Caesar, V. Bankiti, and A. H. Lang, S. Vora, V. E. Liong, Q. Xu, Qiang A. Krishnan, Y.Pan, G. Baldan, O. Beijbom, nuscenes: A multimodal dataset for autonomous driving, In Proceedings of the IEEE/CVF Conference on Computer Vision and Pattern Recognition, pp. 11621-11631, 2020.

\bibitem{semantickitti} J. Behley, M. Garbade, A. Milioto, J. Quenzel, S. Behnke, C. Stachniss, J. Gall, Semantickitti: A dataset for semantic scene understanding of lidar sequences, In Proceedings of the IEEE/CVF International Conference on Computer Vision, pp. 9297-9307, 2019.

\bibitem{rgbal} K. El Madawi, H. Rashed, A. El Sallab, O. Nasr, H. Kamel, S. Yogamani, Rgb and lidar fusion based 3d semantic segmentation for autonomous driving,  IEEE Intelligent Transportation Systems Conference, pp. 7-12, 2019.

\bibitem{pointpainting} S. Vora, A. H. Lang, B. Helou, O. Beijbom, Pointpainting: Sequential fusion for 3d object detection, In Proceedings of the IEEE/CVF Conference on Computer Vision and Pattern Recognition, pp. 4604-4612, 2020.

\bibitem{fusionpainting} S. Xu, D. Zhou, J. Fang, J. Yin, Z. Bin, L. Zhang, FusionPainting: Multimodal fusion with adaptive attention for 3d object detection, IEEE Intelligent Transportation Systems Conference, pp. 4604-4612, 2021.

\bibitem{pmf} Z. Zhuang, R. Li, K. Jia, Q. Wang, Y. Li, M. Tan, Perception-aware multi-sensor fusion for 3d lidar semantic segmentation, In Proceedings of the IEEE/CVF International Conference on Computer Vision, pp. 16280-16290, 2021.

\bibitem{tornadonet} M. Gerdzhev, R. Razani, E. Taghavi, L. Bingbing, Tornado-net: multiview total variation semantic segmentation with diamond inception module, IEEE International Conference on Robotics and Automation, pp. 9543-9549, 2021.

\bibitem{salsanext} T. Cortinhal, G. Tzelepis, E. Erdal Aksoy, SalsaNext: Fast, uncertainty-aware semantic segmentation of LiDAR point clouds, Advances in Visual Computing, pp. 207-222, 2020.

\bibitem{rangenet++} A. Milioto, I. Vizzo, J. Behley, C. Stachniss, Rangenet++: Fast and accurate lidar semantic segmentation, IEEE/RSJ International Conference on Intelligent Robots and Systems, pp. 4213-4220, 2019.

\bibitem{cylinder3d} X. Zhu, H. Zhou, T. Wang, F. Hong, Y. Ma, W. Li, Wei H. Li, D. Lin, Cylindrical and asymmetrical 3d convolution networks for lidar segmentation, In Proceedings of the IEEE/CVF Conference on Computer Vision and Pattern Recognition, pp. 9939-9948, 2021.

\bibitem{af2s3net} R. Cheng, R. Razani, E. Taghavi, E. Li, B. Liu, 2-s3net: Attentive feature fusion with adaptive feature selection for sparse semantic segmentation network, In Proceedings of the IEEE/CVF Conference on Computer Vision and Pattern Recognition, pp. 12547-12556, 2021.


\bibitem{kpconv} H. Thomas, C. R. Qi, J. E. Deschaud, B. Marcotegui, F. Goulette, L. J. Guibas, Kpconv: Flexible and deformable convolution for point clouds, In Proceedings of the IEEE/CVF International Conference on Computer Vision, pp. 6411-6420, 2019.

\bibitem{rpvnet} J. Xu, R. Zhang, J. Dou, Y. Zhu, J. Sun, S. Pu, Rpvnet: A deep and efficient range-point-voxel fusion network for lidar point cloud segmentation, In Proceedings of the IEEE/CVF International Conference on Computer Vision, pp. 16024-16033, 2021.

\bibitem{fcn} J. Long, E. Shelhamer, T. Darrell, Fully convolutional networks for semantic segmentation, In Proceedings of the IEEE/CVF Conference on Computer Vision and Pattern Recognition, pp. 3431-3440, 2015.

\bibitem{mul1} H. Zhao, J. Shi, X. Qi, X. Wang, J. Jia, Pyramid scene parsing network, In Proceedings of the IEEE/CVF Conference on Computer Vision and Pattern Recognition, pp. 2881-2890, 2017.

\bibitem{mul2} H. Zhao, J. Shi, X. Qi, X. Wang, J. Jia, Laplacian pyramid reconstruction and refinement for semantic segmentation, In Proceedings of the European Conference on Computer Vision, pp. 519-534, 2016.

\bibitem{dil1} L. C. Chen, G.  Papandreou, F. Schroff, H. Adam, Rethinking atrous convolution for semantic image segmentation, arXiv preprint arXiv:1706.05587, 2017.

\bibitem{dil2}  L. C. Chen Y. Zhu, G. Papandreou, F. Schroff, H. Adam, Encoder-decoder with atrous separable convolution for semantic image segmentation, In Proceedings of the European Conference on Computer Vision, pp. 801-818, 2018.

\bibitem{att1} L. C. Chen, Y. Yang, J. Wang, W. Xu, A. L. Yuille, Attention to scale: Scale-aware semantic image segmentation, In Proceedings of the IEEE/CVF Conference on Computer Vision and Pattern Recognition, pp. 3640-3649, 2016.

\bibitem{att2} Q. Huang, C. Xia, C. Wu, S. Li, Y. Wang, Y. Song, C-C. J Kuo, Semantic segmentation with reverse attention, arXiv preprint arXiv:1707.06426n, 2017.

\bibitem{att3} J. Fu, J. Liu, H. Tian, Y. Li, Y. Bao, Z. Fang, H. Lu, Dual attention network for scene segmentation, In Proceedings of the IEEE/CVF Conference on Computer Vision and Pattern Recognition, pp. 3146-3154, 2019.

\bibitem{segformer} E. Xie, W. Wang, Z. Yu, A. Anandkumar, J. M. Alvarez, P. Luo, Ping, SegFormer: Simple and efficient design for semantic segmentation with transformers, Advances in Neural Information Processing Systems, vol. 34, pp. 12077-12090, 2021.

\bibitem{maskformer} B. Cheng, A. Schwing, A. Kirillov, Per-pixel classification is not all you need for semantic segmentation, Advances in Neural Information Processing Systems, vol. 34, pp. 17864-17875, 2021.

\bibitem{pointnet} C. R. Qi, H. Su, K. Mo, L. J. Guibas, Pointnet: Deep learning on point sets for 3d classification and segmentation, In Proceedings of the IEEE/CVF Conference on Computer Vision and Pattern Recognition, pp. 652-660, 2017.

\bibitem{pointnet++} C. R. Qi, L. Yi, H. Su, L. J. Guibas, Pointnet++: Deep hierarchical feature learning on point sets in a metric space, Advances in Neural Information Processing Systems, vol. 30, 2017.

\bibitem{pointwise} B. S. Hua, M. K. Tran, S. K. Yeung, Pointwise convolutional neural networks, In Proceedings of the IEEE/CVF Conference on Computer Vision and Pattern Recognition, pp. 984-993, 2018.

\bibitem{sscn} B. Graham, M. Engelcke, L. Van Der Maaten, 3d semantic segmentation with submanifold sparse convolutional networks, In Proceedings of the IEEE/CVF Conference on Computer Vision and Pattern Recognition, pp. 9224-9232, 2018.

\bibitem{salsanet} E. E. Aksoy, S. Baci, S. Cavdar, Salsanet: Fast road and vehicle segmentation in lidar point clouds for autonomous driving, IEEE intelligent vehicles symposium, 2020.

\bibitem{bev} C. Zhang, W. Luo, R. Urtasun, Efficient convolutions for real-time semantic segmentation of 3d point clouds, International Conference on 3D Vision, pp. 399-408, 2018.

\bibitem{polarnet} Y. Zhang, Z. Zhou, P. David, X. Yue, Z. Xi, B. Gong, H. Foroosh, Polarnet: An improved grid representation for online lidar point clouds semantic segmentation, In Proceedings of the IEEE/CVF Conference on Computer Vision and Pattern Recognition, pp. 9601-9610, 2020.

\bibitem{amvnet} V. E. Liong, T. N. T. Nguyen,  S. Widjaja, D. Sharma, Z. J. Chong, Amvnet: Assertion-based multi-view fusion network for lidar semantic segmentation, arXiv preprint arXiv:2012.04934, 2020.

\bibitem{pvcnn} Z. Liu, H. Tang, Y. Lin, S. Han, Point-voxel cnn for efficient 3d deep learning, Advances in Neural Information Processing Systems, vol. 32, 2019.

\bibitem{2dpass} X. Yan, J. Gao, C. Zheng, C. Zheng, R. Zhang, S. Cui, Z. Li, 2dpass: 2d priors assisted semantic segmentation on lidar point clouds, In European Conference on Computer Vision, pp. 677-695, 2022.

\bibitem{fuseseg} G. Krispel, M. Opitz, G. Waltner, H. Possegger, H. Bischof, Fuseseg: Lidar point cloud segmentation fusing multi-modal data, In Proceedings of the IEEE/CVF Winter Conference on Applications of Computer Vision, pp. 1874-1883, 2020.

\bibitem{kd} G. Hinton, O. Vinyals, J. Dean, Distilling the knowledge in a neural network, arXiv preprint arXiv:1503.02531, 2015.

\bibitem{kd2} J. Gou, B. Yu, S. J. Maybank, D. Tao, Knowledge distillation: A survey, International Journal of Computer Vision, vol. 129, pp. 1789-1819, 2021.

\bibitem{kd3} J. Xie, B. Shuai, J. F. Hu, J. Lin, W. S. Zheng, Improving fast segmentation with teacher-student learning, arXiv preprint arXiv:1810.08476, 2018.

\bibitem{kd4} Y. Wang, W. Zhou, T. Jiang, X. Bai, Xiang Y. Xu, Intra-class feature variation distillation for semantic segmentation, In European Conference on Computer Vision, pp. 346-362, 2020.

\bibitem{kd5} C. Shu, Y. Liu, J. Gao, Z. Yan, C. Shen, Channel-wise knowledge distillation for dense prediction, In Proceedings of the IEEE/CVF International Conference on Computer Vision, pp. 5311-5320, 2021.

\bibitem{kd6} T. He, C. Shen, Z. Tian, D. Gong, C. Sun, Y. Yan, Knowledge adaptation for efficient semantic segmentation, In Proceedings of the IEEE/CVF Conference on Computer Vision and Pattern Recognition, pp. 578-587, 2019.

\bibitem{kd7} Y. Hou, Z. Ma, C. Liu, T. W. Hui, Tak-Wai C. C. Loy, Inter-region affinity distillation for road marking segmentation, In Proceedings of the IEEE/CVF Conference on Computer Vision and Pattern Recognition, pp. 12486-12495, 2020.

\bibitem{kd8} Y.  Liu, K. Chen, C. Liu, Z. Qin, Z. Luo, J. Wang, Structured knowledge distillation for semantic segmentation, In Proceedings of the IEEE/CVF Conference on Computer Vision and Pattern Recognition, pp. 2604-2613, 2019.

\bibitem{pvkd} Y. Hou, X. Zhu, Y. Ma, C. C. Loy, Y. Li, Point-to-Voxel Knowledge Distillation for LiDAR Semantic Segmentation, In Proceedings of the IEEE/CVF Conference on Computer Vision and Pattern Recognition, pp. 8479-8488, 2022.

\bibitem{efficientlps} K. Sirohi, R. Mohan, D. B{\"u}scher, W. Burgard, A. Valada, Efficientlps: Efficient lidar panoptic segmentation, IEEE Transactions on Robotics, 2021.

\bibitem{cbam} S. Woo, J. Park, J. Y. Lee, I. So Kweon, Cbam: Convolutional block attention module, In European Conference on Computer Vision, pp. 3-19, 2018.

\bibitem{deepfusion} Y. Li, A. W. Yu, T. Meng, B. Caine, J. Ngiam, D. Peng, J. Shen, Y. Lu, D. Zhou, Q. V. Le, Deepfusion: Lidar-camera deep fusion for multi-modal 3d object detection, In Proceedings of the IEEE/CVF Conference on Computer Vision and Pattern Recognition, pp. 17182-17191, 2022.

\bibitem{pvtv2} W. Wang, E. Xie, X. Li, D. P. Fan, K. Song, D. Liang, T. Lu, P. Luo, L. Shao, Pvt v2: Improved baselines with pyramid vision transformer, Computational Visual Media, vol. 8, no. 3, pp. 415-424, 2022.

\bibitem{pytorch} S. Imambi, K. B. Prakash, G. R. Kanagachidambaresan, PyTorch, Programming with TensorFlow, vol. 8, no. 3, pp. 87-104, 2021.

\bibitem{kittistep} M. Weber, J. Xie, M. Collins, Y. Zhu, P. Voigtlaender, H. Adam, B. Green, A. Geiger, B. Leibe, D. Cremers, Step: Segmenting and tracking every pixel, arXiv preprint arXiv:2102.11859, vol. 8, no. 3, pp. 87-104, 2021.

\bibitem{cosine} I. Loshchilov, F. Hutter, Sgdr: Stochastic gradient descent with warm restarts, arXiv preprint arXiv:1608.03983, 2016.

\bibitem{wce} S. Panchapagesan, M. Sun, A. Khare, S. Matsoukas, A. Mandal, B. Hoffmeister, S. Vitaladevuni, Multi-task learning and weighted cross-entropy for DNN-based keyword spotting, Interspeech, vol. 9, pp. 760-764, 2016.

\bibitem{lovasz} M. Berman, A. R. Triki, M. B. Blaschko, The lov{\'a}sz-softmax loss: A tractable surrogate for the optimization of the intersection-over-union measure in neural networks, In Proceedings of the IEEE/CVF Conference on Computer Vision and Pattern Recognition, pp. 4413-4421, 2018.

\bibitem{focal} T. Y. Lin, P. Goyal, R. Girshick, K. He, P. Doll{\'a}r, Focal loss for dense object detection, In Proceedings of the IEEE/CVF International Conference on Computer Vision, pp. 2980-2988, 2017.

\bibitem{squeezesegv3} C. Xu, B. Wu, Z. Wang, W. Zhan, P. Vajda, K. Keutzer, M. Tomizuka, Squeezesegv3: Spatially-adaptive convolution for efficient point-cloud segmentation, In European Conference on Computer Vision, pp. 1-19, 2020.

\bibitem{spvnas} H. Tang, Z. Liu, S. Zhao, Y. Lin, J. Lin, H. Wang, S. Han, Searching efficient 3d architectures with sparse point-voxel convolution, In European Conference on Computer Vision, pp. 685-702, 2020.

\bibitem{resnet} K. He, X. Zhang, S. Ren, J. Sun, Deep residual learning for image recognition, In Proceedings of the IEEE/CVF Conference on Computer Vision and Pattern Recognition, pp. 770-778, 2016.

\bibitem{adam} D. P. Kingma, J. Ba, Adam: A method for stochastic optimization, arXiv preprint arXiv:1412.6980, 2015.


\end{thebibliography}
\end{document}